\documentclass{article}

\usepackage[final]{neurips_2023}

\usepackage[utf8]{inputenc} 
\usepackage[T1]{fontenc}    
\usepackage{hyperref}       
\usepackage{url}            
\usepackage{booktabs}       
\usepackage{amsfonts}       
\usepackage{nicefrac}       
\usepackage{microtype}      
\usepackage[dvipsnames]{xcolor}         
\usepackage{graphicx}
\usepackage{dirtree} 
\usepackage[center]{caption}
\usepackage{placeins}
\usepackage[toc,page]{appendix}

\graphicspath{ {./images/} }

\title{Utilizing Explainability Techniques for Reinforcement Learning Model Assurance}
\author{%
Alexander Tapley\thanks{Corresponding author.} \quad Kyle Gatesman \quad Luis Robaina \quad Brett Bissey \quad Joseph Weissman\\
The MITRE Corporation - \textit{AI Security and Perception} \\
\texttt{\{atapley, kjgatesman, lrobaina, bbissey, weissmanj\}@mitre.org}
}

\begin{document}
\maketitle

\begin{abstract}
\noindent Explainable Reinforcement Learning (XRL) can provide transparency into the decision-making process of a Deep Reinforcement Learning (DRL) model and increase user trust and adoption in real-world use cases. By utilizing XRL techniques, researchers can identify potential vulnerabilities within a trained DRL model prior to deployment, therefore limiting the potential for mission failure or mistakes by the system. This paper introduces the ARLIN (Assured RL Model Interrogation) Toolkit, an open-source Python library that identifies potential vulnerabilities and critical points within trained DRL models through detailed, human-interpretable explainability outputs. To illustrate ARLIN's effectiveness, we provide explainability visualizations and vulnerability analysis for a publicly available DRL model. The open-source code repository is available for download at \texttt{https://github.com/mitre/arlin}.

\end{abstract}

\section{Introduction}
Over the last decade, reinforcement learning has increased in popularity due to its ability to achieve superhuman performance on a variety of classic board \cite{alphago} and video game \cite{atari} environments. This gain in popularity has sparked an interest in using DRL for both decision support and autonomous operation within safety-critical scenarios such as air-to-air combat \cite{air2air}, nuclear power plant optimization \cite{nuclear}, and ballistic missile guidance \cite{ballistic}. These use-cases are considered high-risk as even small mistakes can result in large losses of monetary value, equipment, and life. Before DRL models can safely be deployed within real-world safety critical environments, their associated vulnerabilities need to be identified and understood so effective training enhancements and verification guardrails can be implemented.

In this paper, we present the ARLIN Toolkit, an open-source research library written in Python that provides explainability outputs and vulnerability detection for DRL models, specifically designed to increase model assurance and identify potential points of failure within a trained model. To our knowledge, ARLIN is the first open-sourced Python toolkit focused on utilizing explainability techniques to assure RL models prior to deployment. ARLIN utilizes \textit{matplotlib} \cite{matplotlib} and \textit{networkx} \cite{networkx} to visualize a trained DRL model's decision making process and provide meaningful vulnerability identification and analysis to researchers. The modular library is structured to support custom architectures, algorithms, DRL frameworks, and analytics; and provides a well-documented and tested API for XRL research development and model assurance. The ARLIN repository is available for download at \texttt{https://github.com/mitre/arlin}.

\section{Background and Preliminaries}
\subsection{Reinforcement Learning}
Reinforcement learning is an area of machine learning that focuses on teaching an intelligent agent how to interact within an environment in order to optimize a reward function and achieve a specified goal \cite{Sutton1998}. As the agent interacts with the environment, it receives scaled rewards to indicate good and bad actions. Through trial and error, the agent is able to identify the optimal policy in order to maximize the cumulative reward received and solve the given task.

In DRL, the environment is defined as a \textbf{Markov Decision Process}, MDP, $M = (S, A, P, \rho_0, R, \gamma, T)$, where $S$ is the state space, $A$ is the action space, $P : S \times A \times S \rightarrow [0,1]$ is the state transition probability, $\rho_0 : S \times A \rightarrow [0,1]$ is the initial state probability, $R : S \times A$ is the reward function, $\gamma$ is the discount factor, and $T$ is the maximum episode length. The policy $\pi_\theta : S \times A$ assigns a probability value to an action given a state.

During training, the agent observes the current state of the environment $s_t \in S$ and performs an action $a_t \in A$ according to its policy $\pi_\theta$. The agent then receives a next state $s_t^{\prime} \in S$ and reward $r_t$ from $R$ within the environment. The agent's goal is to find a policy that optimizes $R$. Due to the large state space $S$, neural networks are commonly used as function approximators in DRL tasks. While this helps the agent to generalize to continuous or large state spaces, it reduces transparency into the decision making process of the model.

\subsection{Explainable Reinforcement Learning}
The "black-box" nature of deep neural networks make verifying and understanding their underlying reasoning very difficult. A lot of work has been done in the field of Explainable AI (XAI) in recent years \cite{xai}. However, most of these works focus on supervised learning or unsupervised learning tasks that deal with non-sequential input data which do not directly transfer in the case of DRL due to the sequential nature of the task. The lack of transparency into the decision making process of an DRL model decreases user and public trust and introduces potentially catastrophic unknowns into the model performance, therefore increasing the potential for mission failure.

Explainable RL (XRL) is a field of RL that focuses on increasing DRL model transparency to give users insight into a model's decision making process. The information gained from XRL techniques can help researchers identify \textit{why} agents are making certain decisions and increase user trust in the model. Milani \cite{xrl} buckets current XRL works into 3 main categories: feature importance, learning process and MDP, and policy-level. These categories look into different aspects of the agent's decision making process including the importance of different features on the policy's chosen action, training examples that affect the policy outputs, and overall policy behavior analysis. This interpretability information can be labeled as \textit{local} or \textit{global}, where local explanations focus on interpreting the predictions of a single action at a point in time and global explanations give a holistic view of the policy's behavior overall \cite{xrl}. Our work focuses on the \textit{global} interpretability of an DRL model as we aim to analyze the overarching policy to identify potential critical points that may affect a policy's success.

\subsection{Related Works}
To our knowledge, ARLIN is the first open-sourced Python library focusing on global explainability and vulnerability detection through human-interpretable analysis visualizations. InterestingnessXRL \cite{interestingness} similarly provides explainability outputs for users, but focuses primarily on identifying interesting interactions between the agent and the environment called \textit{highlights} and returns video-samples of the highlights along with analytics about the interaction itself. While vulnerabilities and critical points may be diagnosed as a highlight, this work does not explicitly focus on these areas. While other repositories linked to XRL are publicly available such as \cite{mkrl}, these are providing XRL algorithms themselves as opposed to visualizations and analytics for trained DRL models.

\section{Key Features}
The ARLIN Toolkit provides three main explainability analysis components to users: latent space analysis, datapoint cluster analysis, and semi-aggregated Markov decision process (SAMDP) \cite{samdp} analysis.

\begin{itemize}
    \item \textbf{Latent space analysis} uses dimensionality reduction techniques to generate embeddings from user-specified datapoint metadata and plot them in 2-D space. Additional policy metadata can be overlaid onto the embeddings to visualize the relationship between the policy embeddings and the policy metadata.
    \item \textbf{Datapoint cluster analysis} uses unsupervised clustering methods to cluster datapoints based on user-defined policy metadata and provide analysis on each state cluster. Average metrics for each cluster can be plotted for comparison to identify potential outliers and gain information about what is happening in a specific area of the environment or point in time, such as failure states and critical points. 
    \item \textbf{SAMDP analysis} transforms the identified state clusters into an SAMDP to provide a holistic overview of how the policy moves through the environment over an entire episode. The analysis uses graph theory to identify paths between nodes along with the actions needed to bring the policy from A to B. Paired with the cluster state analysis, users can identify the paths and actions required for a policy to reach an identified failure state or mistake-prone area.
\end{itemize}

\section{Structure and Customizations}
The following is a conceptual overview of the ARLIN library structure along with instructions for adding additional custom components. A practical example usage of the library's methods can be found in Appendix A 1.1.
\subsection{Conceptual Structure}
\begin{figure}[ht]
    \centering
    \includegraphics[width=0.5\textwidth]{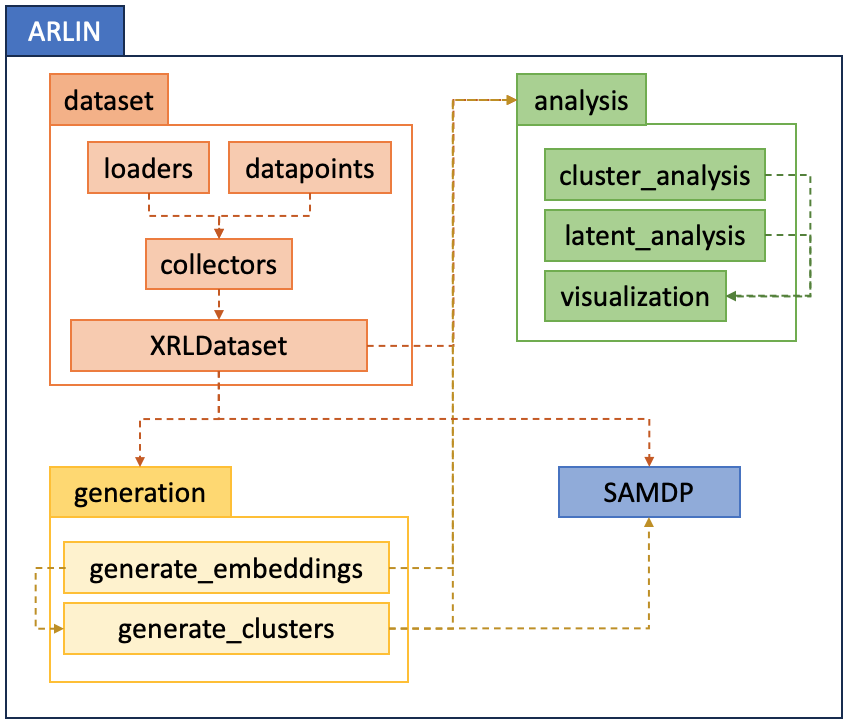}
    \caption{Conceptual structure diagram of the ARLIN library.}
    \label{fig:structure}
\end{figure}
\FloatBarrier

\subsection{Code Structure}
A conceptual diagram of the library structure and relationships between components is shown in Figure \ref{fig:structure}. At a high-level, ARLIN has 4 main components: \textcolor{Melon}{\texttt{dataset}}, \textcolor{Dandelion}{\texttt{generation}}, \textcolor{LimeGreen}{\texttt{analysis}}, and \textcolor{Periwinkle}{\texttt{SAMDP}}. The \textcolor{Melon}{\texttt{dataset}} component is used to create an XRL dataset, a collection of datapoints containing transition data and internal policy metadata collected at every episode step while running a policy within an environment. \textcolor{Dandelion}{\texttt{generation}} uses the XRL dataset to create embeddings and clusters, of which \textcolor{LimeGreen}{\texttt{analysis}} provides meaningful analysis and visualizations. The cluster data and XRL dataset can also be provided to \textcolor{Periwinkle}{\texttt{SAMDP}} to generate and visualize different SAMDP graphs of the agent's policy along with available paths between given clusters.

\colorbox{Melon}{\texttt{dataset}} The \texttt{dataset} directory contains all code necessary for creating an XRL dataset from a trained RL model. \texttt{loaders} handle the loading of a trained model while \texttt{collectors} are responsible for collecting the internal data from the RL model. \texttt{datapoints} outline the specific data that the dataset will be storing. The \texttt{XRLDataset} stores all traditional RL transition data \texttt{(observation, action, reward, done, step)} along with the model-specific metadata (\texttt{Datapoint}) gathered by the \texttt{Collector}. Custom \texttt{loaders}, \texttt{collectors}, and \texttt{datapoints} can be added to load custom models and work with custom architectures and algorithms for the collection of user-defined metadata, as outlined in section \ref{subsection: custom}.

\colorbox{Dandelion}{\texttt{generation}} The \texttt{generation.py} file contains the code necessary for datapoint embedding and cluster generation. Metadata from the XRL dataset chosen by the user is reduced to two dimensions via \texttt{t-SNE} \cite{tsne} to generate latent space embeddings. The datapoints within the XRL dataset are clustered based on user-specified metadata using \texttt{MeanShift} \cite{meanshift} and \texttt{K-Means} \cite{kmeans}. Each cluster represents an area of the policy's latent space where the policy's decision making is affected in similar ways, such as clusters with similar input features or similar output action results.

\colorbox{LimeGreen}{\texttt{analysis}} The \texttt{analysis} directory contains methods for running analysis on both the embeddings as well as the clusters. This includes cluster state representation analysis which analyzes and visualizes the states within the cluster for insight into what states fall into each cluster. The \texttt{visualization} sub-directory contains methods for visualizing the generated analytics using \texttt{matplotlib} \cite{matplotlib}. Example latent analysis and cluster analysis visualizations can be found in Appendices \ref{appendix:latent} and \ref{appendix:cluster}, respectively.

\colorbox{Periwinkle}{samdp} The \texttt{samdp.py} file includes the \texttt{SAMDP} class and associated methods. The \texttt{SAMDP} class is a semi-aggregated Markov decision process representation of the policy within its training environment. The SAMDP methods visualize the connections between clusters as well as available paths and actions required to travel to specific target clusters. Available SAMDP methods and visualizations are attached in Appendix \ref{appendix:samdp}.

\subsection{Custom Component Creation}
\label{subsection: custom}
The modular architecture of ARLIN provides support for user customization with no changes to the main library code. Requirements for creating custom components for common aspects of the library are detailed below:

\textbf{Loaders:} The addition of new loaders does not require any inheritance and can be created as a separate method specific to the model that is being loaded. A custom loader must return a trained model with which a user can run inference within the training environment.

\textbf{Datapoints:} To create a new datapoint, the user must inherit from \texttt{BaseDatapoint} and add any additional metadata that the \texttt{XRLDataset} will be storing for the user-specific use case. A datapoint holds information gathered at a single episode step and can store model-specific internal metadata gathered during the model's decision making process.

\textbf{Collectors:} Custom collectors must inherit from \texttt{BaseDataCollector} and implement the required methods. The collector is responsible for collecting the data needed to fill the datapoint, and therefore is specific to the model architecture as the collector needs to understand where to find the necessary metadata to store.

\textbf{Latent and Cluster Analytics:} To create new analysis visualizations, users can simply create a custom method that produces the wanted metric and return a \texttt{GraphData} object for input into the provided visualization methods.

\section{Usage}
\begin{figure}[ht]
    \centering
    \includegraphics[width=\textwidth]{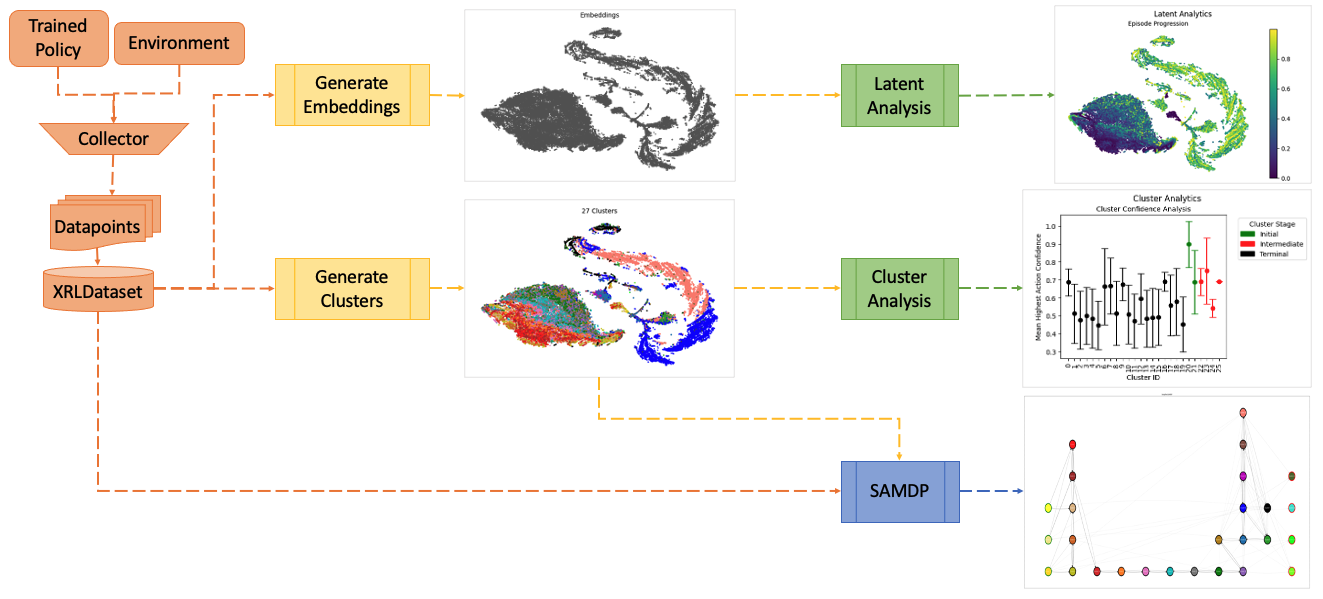}
    \caption{Example workflow to generate vulnerability analysis visualizations using the ARLIN Toolkit.}
    \label{fig:workflow_example}
\end{figure}

ARLIN is designed to provide users with explainability outputs that can be analyzed to identify potential vulnerabilities and critical points within a trained policy. An example workflow for using ARLIN is shown in Figure \ref{fig:workflow_example}. 

\FloatBarrier

To illustrate ARLIN's effectiveness, we provide explainability outputs and corresponding vulnerability analysis for a publicly available DRL model - a model trained using Stable Baselines3 \cite{sb3} with PPO \cite{ppo} on OpenAI gym's Lunarlander-v2 environment \cite{gym}, pulled from Huggingface.com - by following the steps outlined in Figure \ref{fig:workflow_example}. The output visualizations and analysis can be found in Appendix \ref{appendix:latent} (latent analysis), Appendix \ref{appendix:cluster} (cluster analysis), and Appendix \ref{appendix:samdp} (SAMDP analysis).

\section{Discussion and Future Work}
We believe that ARLIN can accelerate research in the XRL field by providing a modular research library with an easy-to-use API for generating explainability visualizations for vulnerability and critical point identification and analysis. This work can be applied to practical use domains such as RL-assisted autonomous vehicle verification and validation and the field of adversarial RL. We hope that the library can expand to include additional analytics, metrics, and visualizations as well as add support for new algorithms and frameworks out of the box through continued author maintenance and community development.

\acksection
The authors thank Walker Dimon and Guido Zarrella for helpful discussions throughout the development process. This work was funded by the 2023 MITRE Independent Research and Development Program's Early Career Research Program.

\clearpage

\bibliographystyle{IEEEtran}
\bibliography{references.bib}

\clearpage

\appendix
\section{Latent Analysis Examples}
\label{appendix:latent}
ARLIN's latent analysis methods make use of the embeddings generated by ARLIN's generation component by overlaying user-defined policy metadata over the generated embeddings to visualize how the metadata relates to location within the embedding space. This information can be helpful when working examining the latent space of a policy. Future work can make use of the latent space to identify similar datapoints or regions, or identify ways to traverse the latent space to reach specific outcomes determined by the metadata, such as actions to take.

\begin{figure}[ht]
    \centering
    \includegraphics[width=\textwidth]{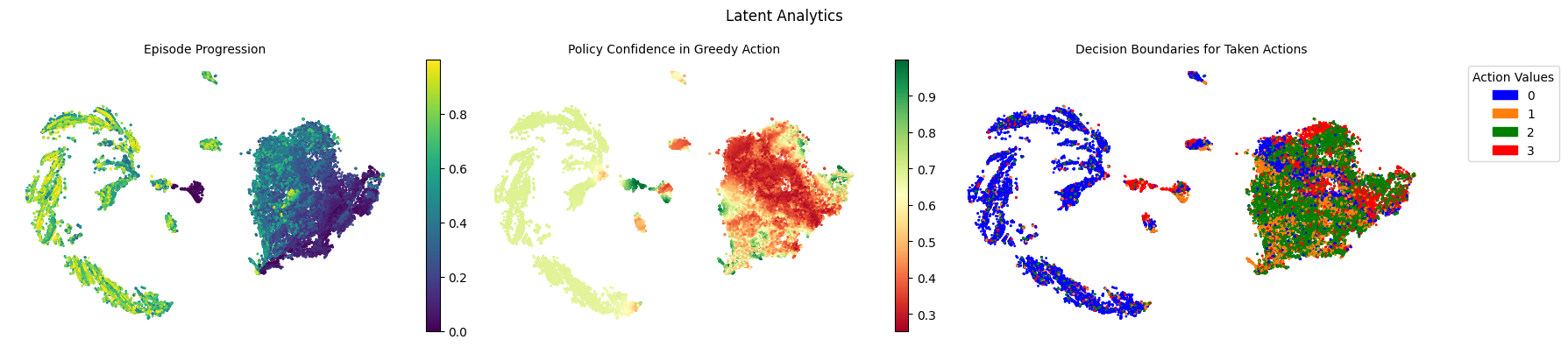}
    \caption{Example outputs from ARLIN's Latent Analysis methods. Graphics show policy metadata overlaid onto embeddings generated from ARLIN's generation component. Overlaid policy metadata overlaid onto embeddings from left to right: episode step that the datapoint was taken at, highest probability from the datapoint's action distribution, action taken at the given datapoint}
    \label{fig:latents}
\end{figure}
\FloatBarrier

\section{Cluster Analysis Examples}
\label{appendix:cluster}
ARLIN's cluster analysis methods make use of the clusters generated by ARLIN's generation component by computing the average values of different policy metadata for each identified cluster. This information gives insight into vulnerable clusters and states within the environment that are reached by the policy. 

The confidence analysis gives insight into how confident the policy is in the action that it is taking at a given point in time. Clusters with low confidence indicate areas of the environment where the policy is not confident in the action that it is taking due to limited training, particularly difficult areas of the environment, or areas where the policy action has no consequence. Clusters with high confidence are areas where the policy is sure of the action it is taking, which can be representative of an easy or very important cluster. A large variance typically represents a cluster where the policy is either very sure or very unsure of its actions, likely resulting in a higher likelihood for mistakes.

The expected return analysis gives insight into both the stage of the episode the cluster is in (early vs late) as well as insight into which states the policy thinks have a higher likelihood for mission success. When looking at initial clusters, a cluster that has a lower expected return is seen as a harder starting position for the policy. When looking at intermediate clusters, clusters with a higher expected return represent "early" stage clusters while "late" stage clusters have a lower expected return.

The reward analysis gives insight into how good the actions taken within the cluster are, represented by the amount of reward received. A higher average reward means the actions are considered better overall. When analyzing terminal clusters, mission failures can be typically be identified by clusters with a large negative reward.

\begin{figure}[ht]
    \centering
    \includegraphics[width=\textwidth]{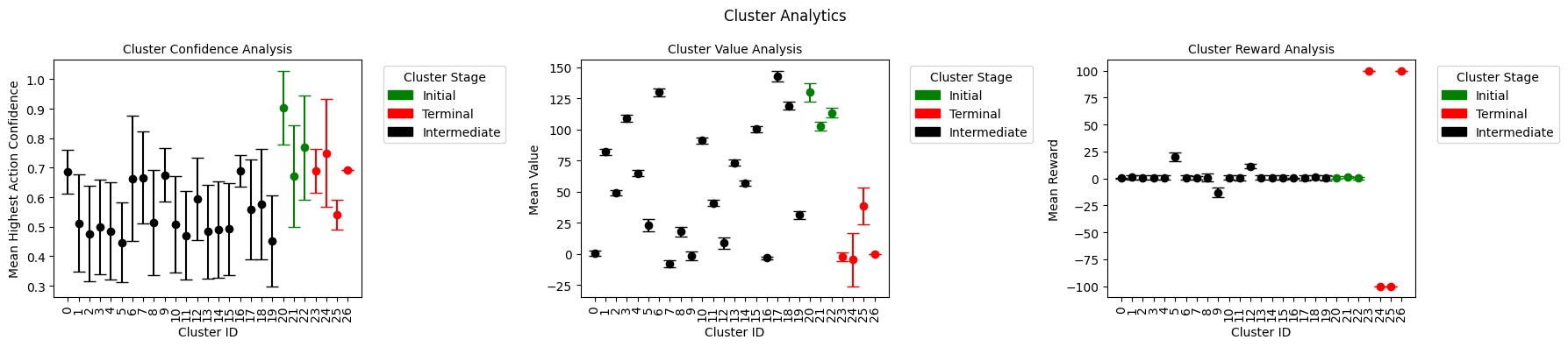}
    \caption{Example outputs from ARLIN's Cluster Analysis methods. Cluster analytics give average metrics for each cluster generated by ARLIN's generation component. From left to right: average greedy action confidence per cluster, average expected return per cluster, average reward per cluster.}
    \label{fig:clusters}
\end{figure}
\FloatBarrier

In Figure \ref{fig:clusters}, we can make a few assumptions about the clusters within our policy. When analyzing initial clusters, Cluster 21 has a low confidence and low expected return, indicating that the cluster is seen as a non-optimal starting position. The policy does not expect to get as much reward overall when starting in Cluster 21 than Cluster 20. When looking at intermediate cluster, Cluster 9 shows a low received reward but high confidence, indicating that it is likely a corrective maneuver that the policy feels is important to take. We can assume this is a late-stage maneuver as well given that the expected return is near 0. For terminal clusters, Cluster 23 has a low expected return and a low received reward, meaning this is likely an expected failure - the policy was expecting a low reward and got a low reward. Cluster 24, however, has a high expected reward but a low received reward, indicating an unexpected failure - the policy was expecting to get more than it received.

\begin{figure}[ht]
    \centering
    \includegraphics[width=\textwidth]{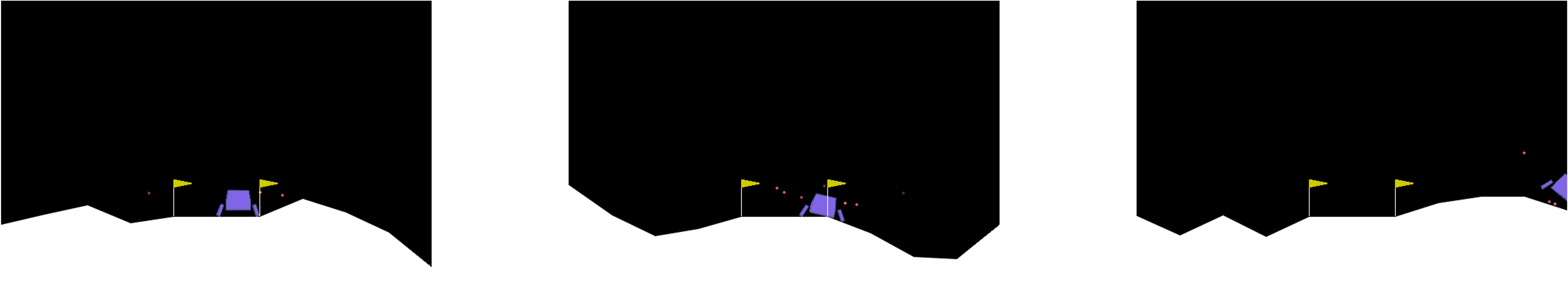}
    \caption{Cluster state analysis from ARLIN showing example images from different clusters. From left to right: Cluster 9 (intermediate), Cluster 23 (terminal), Cluster 24 (terminal).}
    \label{fig:cluster_states}
\end{figure}
\FloatBarrier

As seen in Figure \ref{fig:cluster_states}, our assumptions were correct. Cluster 9 is a late-stage corrective maneuver in which the agent is attempting to move further left to be inside the landing flags. Cluster 23 is an expected failure where the policy lands hard into the ground and crashes, and Cluster 24 is an unexpected failure in which the policy moves off screen, resulting in the end of the episode without a successful landing or a crash.

\section{SAMDP Examples}
\label{appendix:samdp}
\begin{figure}[ht]
    \centering
    \includegraphics[width=0.76\textwidth]{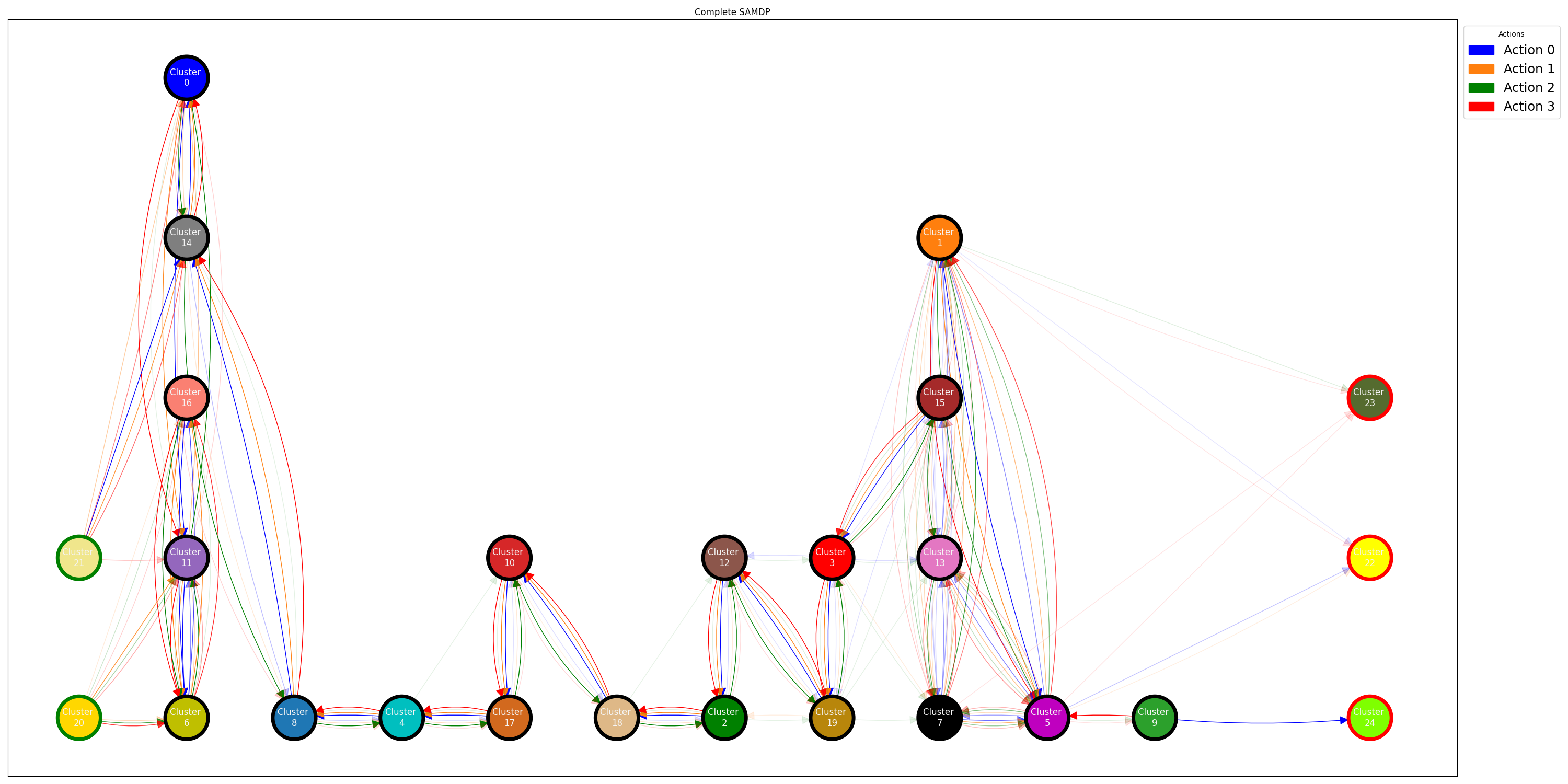}
    \caption{Fully verbose SAMDP showing connections between clusters and the action value that creates the movement.}
    \label{fig:complete}
\end{figure}

\begin{figure}[ht]
    \centering
    \includegraphics[width=0.76\textwidth]{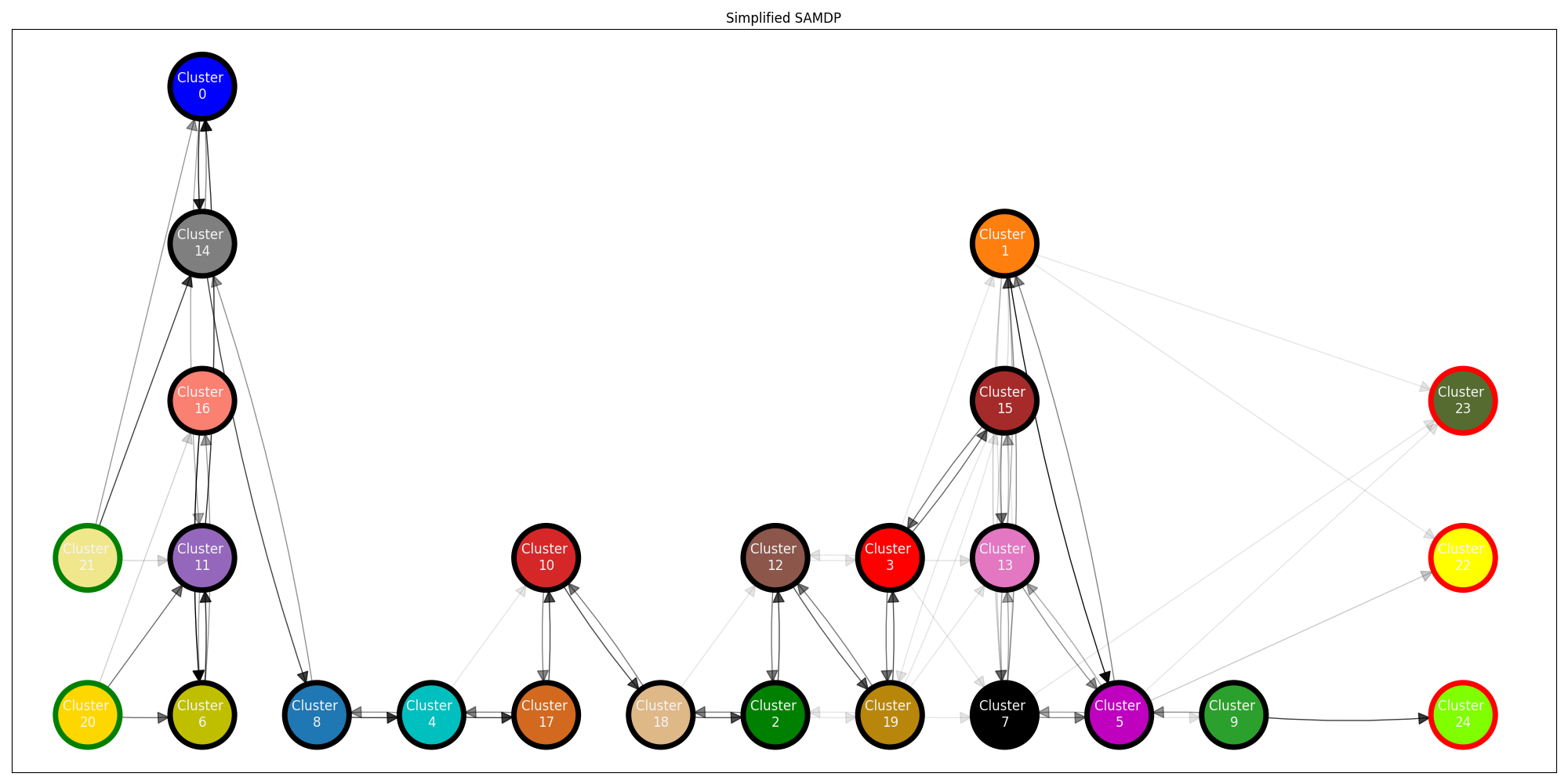}
    \caption{Simplified SAMDP view showing conenctions between clusters regardless of taken action.}
    \label{fig:simplified}
\end{figure}

\begin{figure}[ht]
    \centering
    \includegraphics[width=0.76\textwidth]{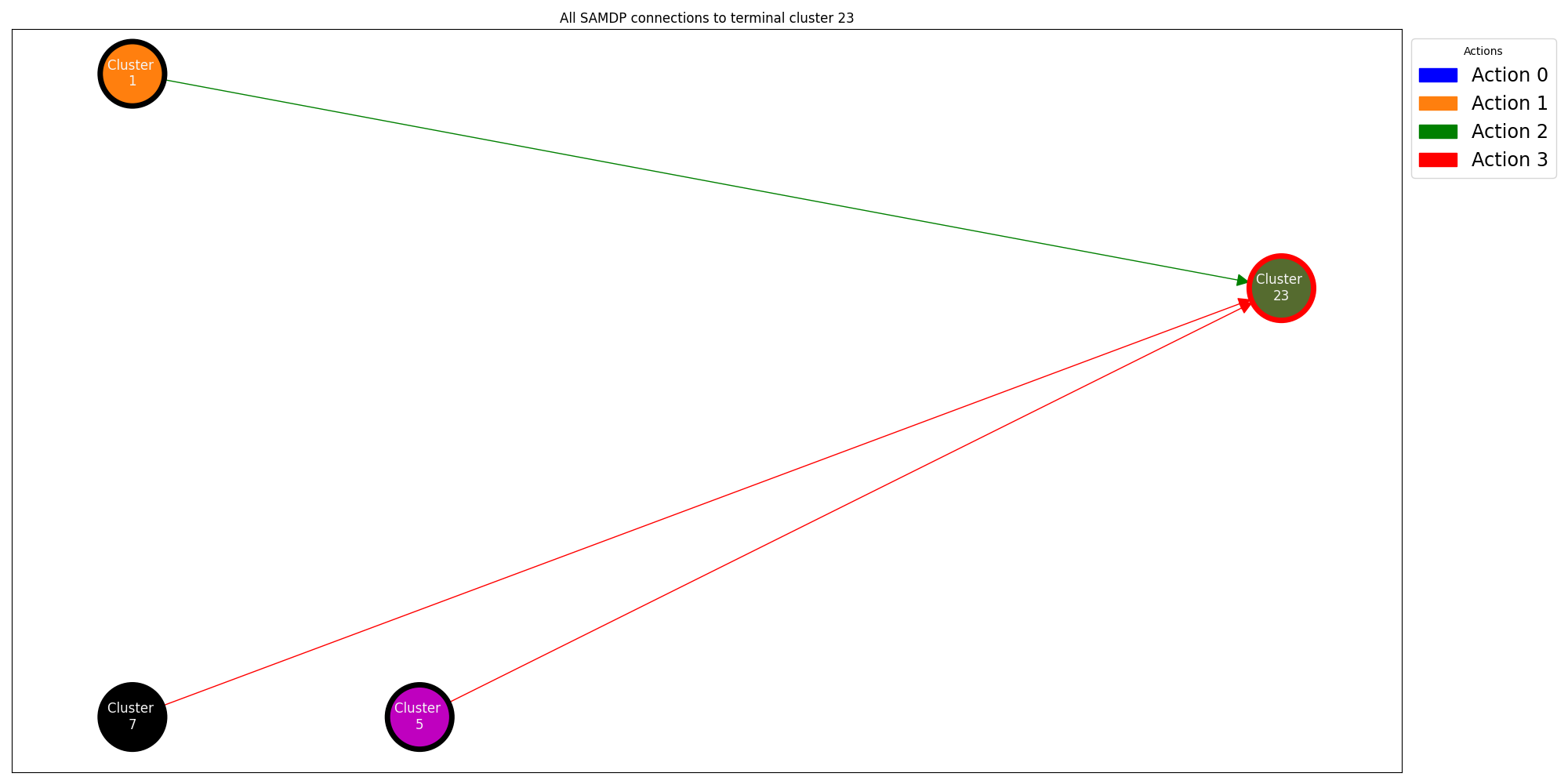}
    \caption{SAMDP view to show movements into all terminal clusters and the actions that bring the agent from a given cluster into the terminal clusters.}
    \label{fig:terminals}
\end{figure}

ARLIN's SAMDP component uses metadata from the XRLDataset along with the generated clusters to generate a semi-aggregated Markov decision process of the policy to show how the policy moves between clusters over the course of an episode. This information is useful in identifying paths between clusters. For vulnerability analysis, this is useful in identifying which actions lead an agent to mission failure, and which actions lead to mission success as well as identifying the critical points where the agent can go either way depending on the actions taken. ARLIN provides a variety of methods in the SAMDP package including holistic views of the entire SAMDP (Figure \ref{fig:complete}), paths between given clusters, or paths leading into a terminal state (Figure \ref{fig:terminals}). All SAMDP methods can provide a full verbose view including the actions necessary for the movement (Figure \ref{fig:complete}), or a simplified view which only shows the connections and not the actions required (Figure \ref{fig:simplified}). Some methods provide the option to only show the most probable connections as well, to avoid connections that are have been taken at least once, but are not likely to be taken by the policy in general.

\FloatBarrier

\FloatBarrier

\FloatBarrier

\end{document}